\PassOptionsToPackage{vietnamese,english}{babel}
\documentclass{article}
\usepackage{makecell} % for additional cell formatting options
\usepackage{array} % Add this line to the preamble
\usepackage{makecell} % Add this line to the preamble
\usepackage{booktabs}
\usepackage{float} 
\usepackage{multirow}
\usepackage{booktabs}
\usepackage{array}
\usepackage{graphicx}
\usepackage{ragged2e}
\usepackage{xcolor}
\usepackage[english]{babel}
\usepackage{subcaption}
\usepackage[letterpaper,top=2cm,bottom=2cm,left=3cm,right=3cm,marginparwidth=1.75cm]{geometry}
\usepackage[utf8]{inputenc}
\usepackage{biblatex} %Imports biblatex package
\usepackage{amsmath}
\usepackage{graphicx}
\usepackage[colorlinks=true, allcolors=blue]{hyperref}
\usepackage{tcolorbox}
\usepackage{enumitem}
\usepackage{authblk}
\usepackage{csquotes}
\usepackage{makecell}
\usepackage{tabularx}

\title{ViBidirectionMT-Eval: Machine Translation for Vietnamese-Chinese and Vietnamese-Lao language pair}
\author[1*]{Hong-Viet Tran}
\author[1]{Minh-Quy Nguyen}
\author[1]{Van-Vinh Nguyen}
\affil[1]{University of Engineering and Technology\\

Vietnam National University, Hanoi, Vietnam

\small Email: \tt{thviet@vnu.edu.vn, minhquy1624@gmail.com, vinhnv@vnu.edu.vn}

\small {*Corresponding author}
}

\begin{document}
\maketitle

\begin{abstract}
% This paper reports the overview of the VLSP 2023 – Machine Translation shared task for Vietnamese-Chinese Machine Translation. This task is hosted at the 10th annual workshop on Vietnamese Language and Speech Processing (VLSP 2023). The goal of Machine Translation shared task is to develop machine translation systems. In the scope of Machine Translation shared task, we focus on Vietnamese, Chinese and Lao news and build a human-annotated dataset. For Vietnamese-Chinese: 300,348 sentences pairs for training data and 1,000 sentences pairs for development and 1,000 sentences pairs test set, collected from Vietnamese news and Chinese. For Vietnamese-Lao: 100,000 sentences pairs for training data and 2,000 sentences pairs for development and 1,000 sentences pairs test set . Participated models are evaluated by translation machines with BLEU [1], SacreBLEU [2] score and ranked in terms of human score with experts in Chinese languages and Lao languages, the most typical evaluation metric for machine translation problem.

This paper presents an results of the VLSP 2022-2023 Machine Translation Shared Tasks, focusing on Vietnamese-Chinese and Vietnamese-Lao machine translation. The tasks were organized as part of the 9th, 10th annual workshop on Vietnamese Language and Speech Processing (VLSP 2022, VLSP 2023). The objective of the shared task was to build machine translation systems, specifically targeting Vietnamese-Chinese and Vietnamese-Lao translation (corresponding to 4 translation directions). The submission were evaluated on 1,000 pairs for testing (news and general domains) using established metrics like BLEU~\cite{Papineni} and SacreBLEU~\cite{Matt:2018}. Additionally, system outputs also were evaluated with human judgment provided by experts in Chinese and Lao languages. These human assessments played a crucial role in ranking the performance of the machine translation models, ensuring a more comprehensive evaluation.
\end{abstract}

\section{Introduction}
\label{sec:intro}
Neural Machine Translation (NMT) has currently obtained state-of-the-art in machine translation systems. However, the translation quality is still a challenge in translation systems. Neural Machine Translation (NMT)~\cite{Cho2014LearningPR, Sutskever:2014, Vaswani:2017} has recently shown impressive results compared to Statistical Machine Translation (SMT)~\cite{Wu:2016, KleinKDSR17}. However, NMT systems still have great challenges~\cite{koehn:2017}. The MT track basically corresponds to a subtitling translation task. The natural translation unit considered by the human translators volunteering for News is indeed the single caption - as defined by the original transcript - which in general does not correspond to a sentence, but to fragments of it that fit the caption space. While translators can look at the context of the single captions, arranging the MT task in this way would make it particularly difficult, especially when word reordering across consecutive captions occurs. For this reason, we preprocessed all the parallel texts to rebuild the original sentences, thus simplifying the MT task. Table \ref{tb:1} provides statistics on in-domain texts supplied for training and evaluation purposes for each MT task. Texts are pre-processed (tokenization, Chinese and Vietnamese segmentation) with the tools used for setting-up baseline systems (see below). For this purpose, the task involved creating a comprehensive dataset with human-annotated translations\footnote{\href{https://huggingface.co/datasets/VLSP2023-MT/ViBidirectionMT-Eval}{https://huggingface.co/datasets/VLSP2023-MT/ViBidirectionMT-Eval}}. 

% In 2022, the shared task on Machine Translation (MT) in the VLSP evaluation campaign marked a successful comeback of Machine Translation in VLSP activities with a handful of teams competing on Chinese-Vietnamese Translation over the News.  Chinese-Vietnamese Machine Translations (Chinese-Vietnamese and Vietnamese-Chinese): The participants need to deal with a scenario in which we do not have much data. Furthermore, since Chinese and Vietnamese can be considered similar languages to some degree (e.g., many Sino Vietnamese words have a 1-to-1 mapping in meaning with their Chinese counterpart), participants could employ unique methods for similar language pairs in this task. 

% In 2023, VLSP evaluation campaign focus on two sub tasks Lao-Vietnamese and Vietnamese-Lao Machine Translation: The participants need to deal with a scenario in which we do not have much data. Furthermore, since Lao and Vietnamese can be considered similar languages to some degree (e.g., many Vietnamese words have a 1-to-1 mapping in meaning with their Lao counterpart), participants could employ unique methods for similar language pairs in this task.

In 2022, a significant milestone was reached for Machine Translation (MT) within the VLSP evaluation campaign, driven by the notable contributions of teams focusing on Chinese–Vietnamese translation through news sources. Despite data scarcity posing a major challenge, participating teams successfully leveraged the linguistic similarities between Chinese and Vietnamese most notably the one-to-one mapping between Sino Vietnamese and Chinese words to develop specialized methods. Moving into 2023, VLSP has shifted its attention to Lao–Vietnamese and Vietnamese–Lao Machine Translation tasks, where limited training data continues to hinder model development. Nonetheless, the substantial similarities between Lao and Vietnamese, including numerous one-to-one lexical mappings, open opportunities to apply specialized techniques for this closely related language pair, even under constrained data conditions.

\section{Training \& Test Data}

In the VLSP 2022 and VLSP 2023 evaluation campaign, we released comprehensive training datasets designed to support Vietnamese-Chinese and Vietnamese-Lao machine translation tasks. These datasets include development and public test sets to facilitate model optimization and evaluation. Specifically, the VLSP 2022 dataset comprises over 300,000 Vietnamese-Chinese bilingual sentence pairs for training, with an additional 1,000 sentences for development and testing. Similarly, the VLSP 2023 dataset, designed for Vietnamese-Lao translation, contains 100,000 bilingual sentence pairs for training, 2,000 for development, and 1,000 for testing.

The provision of development and public test sets allows participants to fine-tune their models before formal evaluation on the secure private test set. Notably, all development, public test, and private test sets are within the same linguistic domain, ensuring consistency in evaluation and benchmarking. SacreBLEU is recommended for model evaluation, as it offers a reliable metric for assessing machine translation accuracy.

The input data is provided in UTF-8 text format, with 1-to-1 aligned bilingual sentence pairs, ensuring precise correspondence throughout training and testing. This approach facilitates standardization and improves the accuracy of machine translation systems, contributing to research and application in automatic translation for Vietnamese in a multilingual context.

Table \ref{tb:1} shows statistics on in-domain texts supplied for training and evaluation purposes for two MT tasks: Vietnamese $\leftrightarrow$ Chinese Machine Translation Systems for VLSP 2022 and Vietnamese $\leftrightarrow$ Lao Machine Translation Systems for VLSP 2023. All parallel texts were tokenized and truncated using sentence piece scripts, and then they are applied to Sennrich’s BPE \cite{Sennrich2015NeuralMT}. For Vietnamese, we only apply Moses’s scripts for tokenization and true-casing.
\section{Evaluation}

The participants to the MT track had to provide the automatic translation of the test sets in text
format. The output had to be case-sensitive, detokenized and had to contain punctuation. The quality of the translations was measured both automatically, against the human translations created
by the open translation project, and via human evaluation (Section \ref{sec:experiment}).

Case sensitive scores were calculated for the three automatic standard metrics BLEU~\cite{Papineni} and
SacreBLEU~\cite{Matt:2018}, as implemented in mteval-v13a.pl and sacrebleu, by calling:

\begin{itemize}
    \item mteval-v13a.pl -c
    \item sacrebleu -t vlsp2022/systems -l zh-vi --echo MTTracks
    \item sacrebleu -t vlsp2023/systems -l lo-vi --echo MTTracks
\end{itemize}

% \begin{table}[H]
% \centering
% \caption{Bilingual training and evaluation corpora statistics.}
% \label{tb:1}
% \begin{tabular}{|c|l|c|c|c|}
% \hline
% \multirow{2}{*}{Task} & \multirow{2}{*}{Dataset} & \multirow{2}{*}{Sent} & \multicolumn{2}{c|}{Tokens} \\ \cline{4-5}
%                       &                          &                       & \multicolumn{1}{l|}{Vi}     & Zh / Lo  \\ \hline
% Vi $\leftrightarrow$ Zh & Train                    & 300,348               & \multicolumn{1}{c|}{43,762} & 141,879 \\ \cline{2-5} 
%                       & Dev                      & 1,000                 & \multicolumn{1}{c|}{2,545}  & 3,796   \\ \cline{2-5} 
%                       & Test                     & 1,000                 & \multicolumn{1}{c|}{2,454}  & 4,078   \\ \hline
% Vi $\leftrightarrow$ Lo & Train                    & 100,000               & \multicolumn{1}{c|}{227,000} & 120,710 \\ \cline{2-5} 
%                       & Dev                      & 2,000                 & \multicolumn{1}{c|}{4,262}  & 1,740   \\ \cline{2-5} 
%                       & Test                     & 1,000                 & \multicolumn{1}{c|}{2,454}  & 4,078   \\ \hline
% \end{tabular}
% \end{table}
% % Please add the following required packages to your document preamble:
% \usepackage{multirow}
\begin{table}[h]
\centering
\caption{Bilingual training and evaluation corpora statistics.}
\label{tb:1}
\begin{tabular}{|c|c|c|ccc|}
\hline
\multirow{2}{*}{\textbf{Task}}           & \multirow{2}{*}{\textbf{Dataset}} & \multirow{2}{*}{\textbf{Sent}} & \multicolumn{3}{c|}{\textbf{Tokens\cite{Sennrich2015NeuralMT}}}                                             \\ \cline{4-6} 
                                         &                                   &                                & \multicolumn{1}{c|}{\textbf{vi}} & \multicolumn{1}{c|}{\textbf{zh}} & \textbf{lo} \\ \hline
\multirow{3}{*}{Vi $\leftrightarrow$ Zh} & Train                             & 300,348                        & \multicolumn{1}{c|}{43,762}      & \multicolumn{1}{c|}{141,879}     & -           \\ \cline{2-6} 
                                         & Dev                               & 1,000                          & \multicolumn{1}{c|}{2,545}       & \multicolumn{1}{c|}{3,796}       & -           \\ \cline{2-6} 
                                         & Test                              & 1,000                          & \multicolumn{1}{c|}{2,454}       & \multicolumn{1}{c|}{4,078}       & -           \\ \hline
\multirow{3}{*}{Vi $\leftrightarrow$ Lo} & Train                             & 100,000                        & \multicolumn{1}{c|}{227,000}     & \multicolumn{1}{c|}{-}           & 120,710     \\ \cline{2-6} 
                                         & Dev                               & 2,000                          & \multicolumn{1}{c|}{4,262}       & \multicolumn{1}{c|}{-}           & 1,740       \\ \cline{2-6} 
                                         & Test                              & 1,000                          & \multicolumn{1}{c|}{2,454}       & \multicolumn{1}{c|}{-}           & 3,126       \\ \hline
\end{tabular}
\end{table}
Detokenized texts were passed, since the two scorers apply an internal tokenizer. Before the
evaluation, Chinese texts were segmented at character level, keeping non-Chinese strings as they are.
In order to allow participants to evaluate their progresses automatically and in identical conditions,
an evaluation server was developed. Participants could submit the translation of any development
set to either a REST Webservice or through a GUI on the web, receiving as output the three scores
BLEU, NIST ~\cite{NIST:2008}, TER ~\cite{TER:2008} and SacreBLEU computed as above. The core of the evaluation server
is a shell script wrapping the mteval scorers. The evaluation server was utilized by the organizers
for the automatic evaluation of the official submissions. After the evaluation period, the evaluation
on test sets was enabled to all participants as well.

% \newpage
% \include{Chapter/Chapter3}

%\section{ Submissions}
\section{System submissions}
\label{syssub}
% \textbf{For VLSP2022-MT:} we received submissions from 05 different sites. The total number of primary runs is 10: 05 Zh-Vi, 05 Vi-Zh. In addition, we were asked to evaluate 10 contrastive runs. The specifics are as follows:
% \begin{itemize}
%     \item Register teams: 25 from domestic and foreign universities and research institutes such as Stanford University, JAIST, HUST, UIT-VNUHCM, VietAI, UET-VNU,... corporations, companies technology such as: Samsung SDS, VinBigData, VCCorp, FTech...
%     \item Data User Agreements: 15 teams
%     \item Submission: 5 teams submitted results including Samsung SDS, VinBigData, JAIST, VCCorp, HUST.
% \end{itemize}
% \textbf{For VLSP2023-MT:} we received submissions from 07 different sites. The total number of primary runs is 10: 07 Lo-Vi, 05 Vi-Lo. In addition, we were asked to evaluate 12 contrastive runs. The specifics are as follows:
% \begin{itemize}
%     \item Register teams: 26 from domestic and foreign universities and research institutes such as HUST, MTA, UET-VNU, HUST DaNang, Fulbright Univerisity Vietnam, UITVNUHCM, US-VNUHCM, PTIT, ... corporations, companies technology such as: Viettel, DMining AT, Innovation Center - VNPT Information Technology Company, Bosch Global Software Technologies Vietnam, DopikAI, Sun-asterisk Inc, BK.AI Center, VCCorp, VND Credit, …
%     \item Data User Agreements: 15 teams
%     \item Submission: 7 teams submitted results including UET-VNU, MTA, Viettel, Bosch Global Software Technologies Vietnam, HUST, Fulbright University Vietnam, US-VNUHCM
% \end{itemize}
In the multilingual machine translation tasks at VLSP 2022 and VLSP 2023, we conducted Vietnamese-to-Chinese and Vietnamese-to-Lao translation tasks, attracting substantial participation from both domestic and international organizations. Specifically:

VLSP 2022 - MT: The machine translation task for Vietnamese-to-Chinese and vice versa had 25 registered teams, including universities such as JAIST and HUST, as well as major corporations like Samsung SDS, VinBigData, and VCCorp. Among these, 5 teams submitted official entries, complete with models for performance evaluation and detailed technical reports.

VLSP 2023 - MT: The Vietnamese to Lao and Lao to Vietnamese translation task attracted 26 registered teams, including institutions and universities such as HUST, MTA, UET-VNU, and technology companies like Viettel and Bosch Global Software Technologies Vietnam. In total, 7 teams submitted official entries for evaluation.

In both machine translation tasks, we selected for each task the three methods that achieved the highest results for each translation task. Each of these methods has technical reports that demonstrate the approach, method content, contribution to the machine translation task, and results achieved. We present each of these methods in each translation task the following section in \ref{syssub1} and \ref{syssub2}
% \newpage
% \include{Chapter/Chapter4}

%\section{Methodology}

\subsection{Vietnamese-Chinese Machine Translation}
\label{syssub1}
% viết thêm đề dẫn chọn 3 phương pháp, lý do lựa chọn, trình bày từng phương pháp: ưu nhược điểm
% In the task of Chinese $\leftrightarrow$ Vietnamese machine translation, we choose 3 methods that achieve the best results for both translation directions. As follows:

In the task of Vietnamese-Chinese bidirectional machine translation, we selected the three most effective approaches to achieve accurate and fluent translations that preserve the original meaning. Each method was carefully evaluated for its translation accuracy and fluency to ensure high-quality, natural output. The teams employed distinct techniques and strategies, including language model fine-tuning and input data optimization, to maximize the quality and naturalness of the translations. The selected methods are as follows: 
\begin{itemize}
    \item {Team 1 (SDS)}: An Efficient Approach for Machine Translation on Low-resource Languages
    \item {Team 2 (VBD-MT)}: VBD-MT Vietnamese-Chinese Bidirectional Translation System
    \item {Team 3 (JNLP)}: An Effective Method using Phrase Mechanism in Neural Machine Translation
\end{itemize}
% \textbf{An Efficient Approach for Machine Translation on Low-resource Languages. A Case Study in Vietnamese-Chinese.
% The main works can be summarized as follows:} 
% \begin{itemize}
%     \item Employing the strong multilingual sequence to sequence the pre-trained language model mBART for the machine translation task.
%     \item Proposing data selection and data synthesis techniques from the monolingual corpus that helps improve machine translation system.
% \end{itemize}

\subsubsection{An Efficient Approach for Machine Translation on Low-resource Languages. }

The team proposes leveraging data synthesis as a technique to augment the training set for low-resource language pairs, particularly Vietnamese-Chinese. To accomplish this, the mBART-50 ~\cite{Tang2020MultilingualTW} machine translation system is first fine-tuned with existing bilingual data. It is then employed to translate from the target language back into the source language, effectively generating a synthetic bilingual dataset. This newly synthesized dataset is subsequently merged with authentic bilingual data, providing a more comprehensive training set for the final model.

In constructing the final translation model, the team follows a systematic approach involving three key steps: (1) Training a Vietnamese-English translation model with mBART-50; (2) Enhancing the dataset by generating additional bilingual data through selected sentences extracted from the monolingual dataset; (3) Fine-tuning the model using this expanded bilingual dataset. The VLSP 2022 dataset, which includes 300,000 bilingual sentence pairs and extensive, cleaned monolingual corpora, is employed to ensure that the input data remains of high quality throughout the training process.

For low-resource language pairs, the team applies the TF-IDF selection technique to identify and extract significant sentences from a large monolingual dataset containing 25 million Vietnamese and 19 million Chinese sentences. The resulting dataset, a synthesized bilingual corpus, is then combined with the original bilingual data to enhance the accuracy and robustness of the final translation model.

The utilization of mBART-50 in this study capitalizes on its multilingual translation capabilities, achieved through denoising training. By supporting up to 50 languages and subsequently fine-tuning the model with VLSP data, the research team successfully developed high-quality multilingual machine translation models specifically tailored to Vietnamese and Chinese, thereby enhancing translation performance for these low-resource languages.

\begin{figure}[H]
    \centering
    \includegraphics[width=16cm, height=5cm]{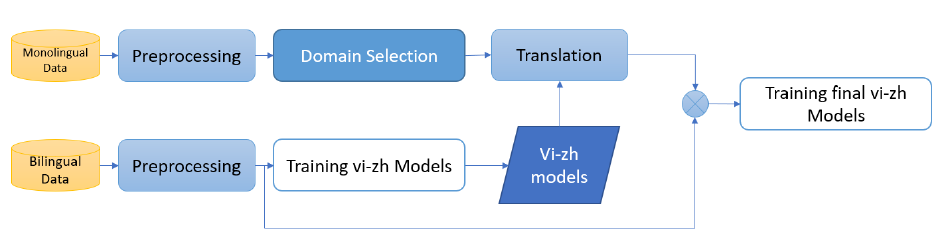}
    \caption {Flow of data processing and model training }
    \label{img1:flow1}
\end{figure}

The system depicted in the figure \ref{img1:flow1} illustrates the training process of a machine translation model for low-resource language pairs, particularly Vietnamese and Chinese. First, both bilingual and monolingual data are processed, and key sentences are selected using the TF-IDF method to ensure domain relevance and maintain high input quality. Subsequently, the mBART-50 model is fine-tuned on the existing bilingual data, then employed to perform back-translation from the target language to the source language, thereby generating a synthetic bilingual dataset. Finally, the authentic bilingual dataset is merged with this synthetic dataset to train the final model, ultimately enhancing both the accuracy and stability of the translation system for Vietnamese and Chinese.

\subsubsection{ VBD-MT Vietnamese-Chinese Bidirectional Translation System }
Baseline system is constructed using the robust Transformer model, which is fine-tuned with mBART-25, a model pre-trained on 25 languages, including both Chinese and Vietnamese. For text processing, we implement the SentencePiece tool to handle tokenization and vocabulary filtering, reducing the vocabulary size from an initial 250K to 67K tokens. This reduction aligns with the limited GPU resources available, allowing for efficient training without the need for high-performance server infrastructure.

To further enhance the dataset, the team apply back-translation using a top-k sampling technique, selecting the top 5 highest-scoring outputs to generate diverse synthetic data. This method yields 211K back-translated sentence pairs from Chinese to Vietnamese (Zh-Vi) and 403K pairs from Vietnamese to Chinese (Vi-Zh). The synthetic data is then combined with authentic bilingual data, effectively expanding the training set and boosting model performance.

The system also employs an ensemble method, achieved by averaging the model weights from the last N checkpoints, where N is optimally set to 5. This ensembling approach significantly enhances accuracy, particularly when used alongside the back-translation data.

To address potential translation errors in numeric and date-time values, the team introduce a post-processing step with customized patterns designed for these data types. Although this post-processing does not directly increase the BLEU score, it improves translation quality by ensuring that critical values, such as those related to people and currency, are accurately translated.

\begin{figure}[h]
    \centering
    \includegraphics[width=14cm, height=5cm]{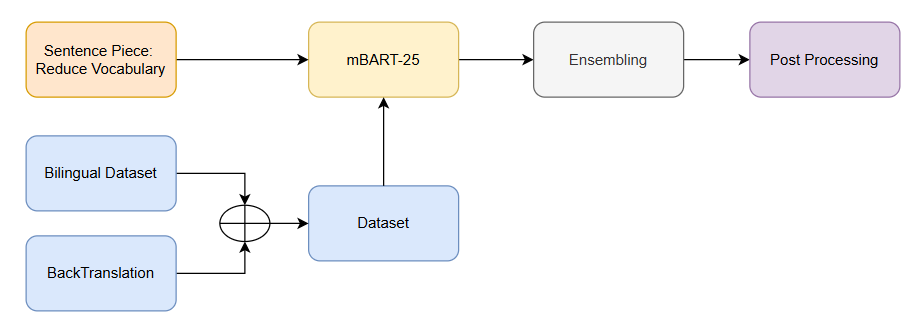}
    \caption { System flow machine translation }
    \label{img2:flow2}
\end{figure}

The system~\ref{img2:flow2} finetunes a Transformer model in conjunction with mBART-25 and employs SentencePiece to reduce the vocabulary size in compliance with GPU resource constraints. The team applies back-translation to generate additional bilingual data, which is then merged with the original dataset to expand the training corpus. The system also utilizes an ensemble technique to enhance accuracy. Finally, a post-processing step is added to correct errors concerning numerical data and dates.

After evaluating various models, the team selected Fairseq~\cite{ott-etal-2019-fairseq} for the baseline system, as it demonstrated superior performance on the public test set. For Chinese-Vietnamese translation, the model achieved a BLEU score of 38.0, which improved to 38.8 with the inclusion of back-translation; for Vietnamese-Chinese translation, the BLEU score increased from 37.8 to 38.0 with the addition of both back-translation and ensembling.

Final system submission for the shared task integrates baseline modeling, back-translation, ensembling, and post-processing. The post-processing step, focused on manually correcting numeric and date-time values, ensures a more accurate and higher-quality translation output for critical data, providing a well-rounded, effective solution.

\subsubsection{ An Effective Method using Phrase Mechanism in Neural Machine Translation }

This approarch developed PhraseTransformer, a model based on the Transformer architecture that incorporates phrase-based attention mechanisms to improve machine translation performance. Unlike prior models, PhraseTransformer eliminates the need for external syntactic tree information, making it more efficient and lightweight compared to other phrase-level attention models. The core concept behind PhraseTransformer is to enhance word representations by leveraging local context and capturing dependencies between phrases within a sentence, enabling more nuanced translation outputs.

In the preprocessing stage, they utilized Byte-Pair Encoding (BPE) to address out-of-vocabulary issues by breaking words down into sub-tokens. For Vietnamese, this work performed 4,000 BPE operations, while for Chinese, which lacks inherent word spacing, they applied 16,000 operations. For Chinese text, the BPE segmentation module treats the entire raw sentence as a single word segment, ensuring effective sub-tokenization even without spacing between characters.

To evaluate PhraseTransformer’s performance against the original Transformer model, they trained both models on the Chinese-Vietnamese bilingual dataset provided by VLSP 2022, without any supplementary external data or pretrained models. Both models were tested under identical configurations, and they averaged the weights from the last 5 checkpoints to produce the final model used for translation testing.

The experimental results reveal that PhraseTransformer consistently outperforms the original Transformer across various n-gram sizes, underscoring the effectiveness of its phrase-based attention mechanism in capturing sentence meaning. Furthermore, PhraseTransformer’s adaptability extends beyond translation tasks to other languages and NLP applications, as it operates independently of external syntactic tree information, making it a versatile tool for diverse linguistic challenges.

\begin{figure}[H]
    \centering
    \includegraphics[width=13cm, height=6cm]{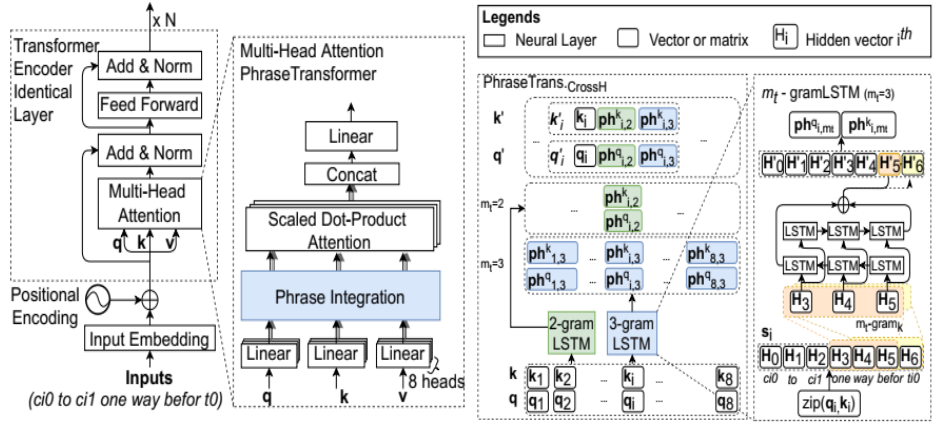}
    \caption{Overview of PhraseTransformer (CrossH) using $n$-gram LSTM in MultiHead layer. In this case, the phrase representations are built with gram\_size = \{2, 3\}, 2-gram, 3-gram models apply to all 8 heads.}
    \label{fig3:flow3}
\end{figure}

The PhraseTransformer (CrossH) system depicted in the figure~\ref{fig3:flow3} is a variant of the Transformer architecture that incorporates a phrase-based attention mechanism to enhance machine translation quality. This model employs n-gram LSTM within multi-head attention to capture local context and inter-phrase relationships, thereby obviating the need for external syntactic tree information.

\subsection{Vietnamese-Lao Machine Translation}
\label{syssub2}

% viết thêm đề dẫn chọn 3 phương pháp, lý do lựa chọn, trình bày từng phương pháp: ưu nhược điểm
% The Vietnamese $\leftrightarrow$ Lao translation task in both directions was carried out using methods based on transformer architectures or sequence-to-sequence models. Both approaches have their advantages in delivering high-quality model outputs. The top-performing teams in the Vietnamese-Lao translation task all utilized pretrained models with transformer architectures.

In the Vietnamese-Lao bidirectional machine translation task, high performance has been achieved using methods based on the Transformer architecture, a leading approach in machine learning. These methods exploit the Transformer’s capacity to deliver high-quality translations by efficiently managing complex word sequences and capturing semantic relationships between words within sentences. Notably, by fine-tuning Transformer-based pretrained models, we can tailor the system to better handle Vietnamese-Lao bilingual data, thereby improving translation accuracy and naturalness. This fine-tuning enhances the system’s precision while also increasing its capacity to capture contextual meaning and accurately reproduce the unique grammatical structures of both languages, achieving high standards in bidirectional  machine translation quality.

In the task of Vietnamese-Lao bidirectional  machine translation, we selected the three most effective approaches, specifically as follows:

\begin{itemize}
    \item {Team 1 (BlueSky)}: A Transformer-Based Model for Lao-Vietnamese Machine Translation
    \item {Team 2 (MTA\_AI)}: Vietnamese-Lao Bidirectional Translation System
    \item {Team 3 (BGSV AI)}: A Sequence-to-Sequence Model for Lao-Vietnamese Machine Translation
\end{itemize}

\subsubsection{ A Transformer-Based Model for Lao-Vietnamese Machine Translation }

Blue\_Sky leverages a pretrained mBART model~\cite{Tang2020MultilingualTW}%~\cite{lin-etal-2022-laoplm} 
initially trained on extensive monolingual datasets in both Vietnamese and Lao. The model is subsequently fine-tuned using a bilingual dataset from VLSP, which enhances its translation accuracy and fluency for both languages. This approach integrates diverse linguistic data from monolingual sources, allowing the model to capture complex grammatical and syntactical structures unique to Vietnamese and Lao, providing a strong base for the fine-tuning phase.

For machine translation tasks, the Transformer WMT~\cite{ott-etal-2018-scaling} en-de big model was employed. This model utilizes an Encoder-Decoder architecture, where the Encoder processes the source sentence to gather context, and the Decoder generates the target sentence sequentially, one word at a time. The model leverages the Transformer’s powerful self-attention mechanism to optimize translation accuracy while maintaining semantic consistency.

To further assess the capabilities of the large language model, this approach adapted mBART for Lao, as the original version of mBART does not support this language. The adaptation involved training mBART on monolingual Vietnamese and Lao datasets to extend its language support, followed by fine-tuning on the bilingual dataset provided by VLSP. This adaptation ensures the model’s effectiveness in machine translation between Vietnamese and Lao.

This method also employed SentencePiece~\cite{kudo-richardson-2018-sentencepiece} for tokenizing Vietnamese and Lao text, setting a vocabulary size of 20,000 tokens. The training dataset consisted of 100,000 sentence pairs, with a test set of 2,000 pairs from VLSP used for evaluation. Additionally, the model was pretrained on a large monolingual dataset 1.8GB of Vietnamese text and 1GB of Lao text—laying a strong foundation for fine-tuning. However, the fine-tuned mBART model performed below the Transformer WMT en-de big model in terms of overall translation accuracy.

\begin{figure}[H]
    \centering
    \includegraphics[width=13cm, height=3.5cm]{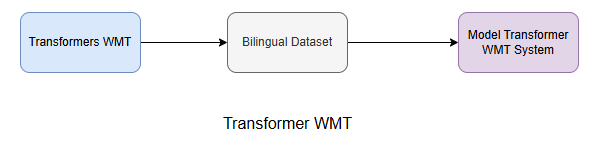}
    \caption {Training phases of mBART and Transformer WMT models.}
    \label{fig4:flow4}
\end{figure}

System~\ref{fig4:flow4} utilizes the Transformer WMT, a conventional encoder-decoder architecture that leverages the self-attention mechanism to optimize machine translation on bilingual datasets.

\subsubsection{ Vietnamese-Lao Bidirectional Translation System }

The MTA\_AI team utilized the M2M-100 418M ~\cite{fan2020englishcentric} and mt5-small \cite{xue-etal-2021-mt5} models to fine-tune a translation system for Vietnamese and Lao, both of which are low-resource languages with limited pre-trained model support. After an extensive survey of available multilingual models, they determined that m2m\_100~\cite{fan2020englishcentric} and mT5 ~\cite{Raffel:2020, xue2021mt5} were particularly well-suited for this project. These models are capable of translating multiple language pairs, including Vietnamese and Lao, making them ideal choices for enhancing translation quality between these languages.

The M2M100-418M model is a multilingual encoder-decoder designed for many-to-many translation, supporting direct translation between numerous languages without needing a pivot language. The mT5-small model, a compact version of T5 with a multilingual capability, was pre-trained on the Common Crawl dataset, covering 101 languages and comprising 300 million parameters. The model is fine-tuned using the Adam optimizer~\cite{kingma2017adam}. This combination allows both comprehensive language support and computational efficiency.

In this approach, MTA\_AI team applied back-translation using Google Translate to convert monolingual sentences into bilingual data, thereby creating a synthetic dataset. This method generated 1.5 million sentence pairs for both Vietnamese-to-Lao (Vi-Lo) and Lao-to-Vietnamese (Lo-Vi) translations, substantially expanding our training data.

To investigate the impact of large-scale data on model performance, they trained the m2m100-418M model with a total of 3 million back-translated monolingual sentences. The results demonstrated a significant enhancement in translation accuracy, affirming the positive influence of large-scale data on the effectiveness of machine translation systems for low-resource languages.

\begin{figure}[H]
    \centering
    \includegraphics[width=13cm, height=6cm]{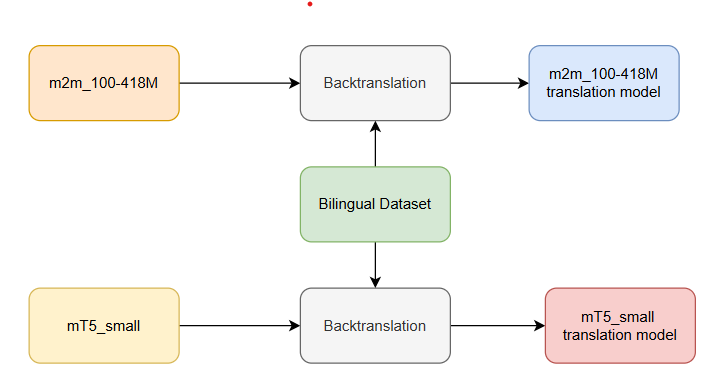}
    \caption {Training phases of mT5\_small and m2m\_100-418M models.}
    \label{fig5:flow5}
\end{figure}

The Fig~\ref{fig5:flow5} depicted in the figure illustrates how the team employs the \texttt{m2m\_100-418M} and \texttt{mT5\_small} models to train machine translation between Vietnamese and Lao using the back-translation method. They generate synthetic bilingual data from monolingual sentences, thereby expanding the training dataset. This approach significantly enhances the translation quality for these two low-resource languages.

\subsubsection{ A Sequence-to-Sequence Model for
Lao-Vietnamese Machine Translation }

%\subsubsection*{The table below describes custom Sequence-to-Sequence Model architecture for Lao-Vietnamese Machine Translation}
% The BGSV\_AI team utilized the T5 model as the base and developed a custom tokenizer from scratch using the SentencePiece technique. They integrated this tokenizer with their custom T5 architecture for the task. Detailed descriptions of the methods, data, and training processes are provided in the table below.
The BGSV\_AI team employed a sequence-to-sequence approach, utilizing large language models to tackle the machine translation challenge in the shared-task competition. During the evaluation phase, they observed that existing models did not support both Vietnamese and Lao simultaneously. Consequently, they developed a unique tokenizer using the SentencePiece technique~\cite{kudo-richardson-2018-sentencepiece} to generate a tailored vocabulary set suited to both languages. They then customized and trained the T5 model ~\cite{Raffel:2020} from scratch, specifically for this machine translation task.

Pre-processing proved essential in enhancing both translation quality and efficiency. This stage involved cleaning the dataset to remove noise, standardizing formats (such as dates and numbers) for uniformity, and tokenizer the text into smaller units. Given their limited familiarity with the Lao language, they applied only fundamental pre-processing techniques, which included the removal of irrelevant characters and symbols.

In the experiment, this approach focused on optimizing the tokenizer to improve sentence comprehension while managing the vocabulary size effectively. To achieve this, BGSV\_AI team sets a token length of 90 for Lao and 150 for Vietnamese, aiming for an optimal balance between computational efficiency and language understanding. These customized token lengths were carefully tailored to the linguistic characteristics of each language, thereby maximizing the performance of their machine translation system.

\begin{figure}[H]
    \centering
    \includegraphics[width=14cm, height=6cm]{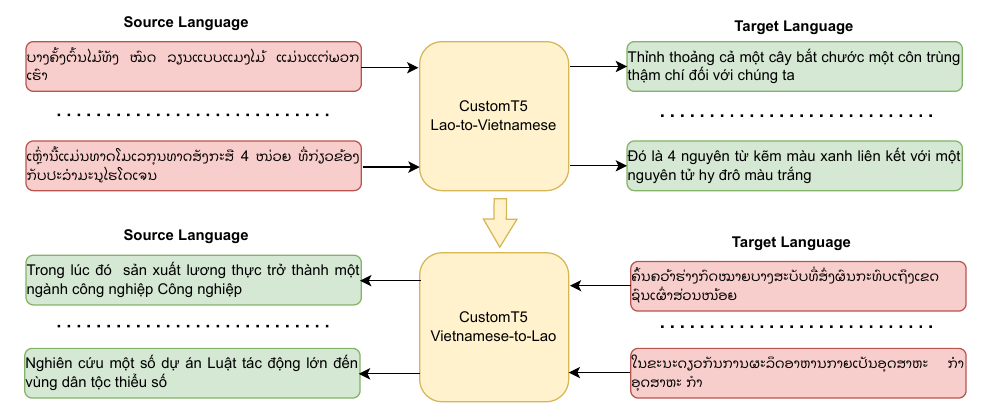}
    \caption {Overview of their proposed machine translation framework, which includes Lao-to-Vietnamese and Vietnamese-to-Lao directions}
    \label{fig6:flow6}
\end{figure}

The Fig \ref{fig6:flow6} below describes custom Sequence-to-Sequence Model architecture for Lao-Vietnamese Machine Translation and an example of translating from Vietnamese to Lao.

% \begin{center}
% %\caption{Sequence-to-Sequence Model architecture for Lao-Vietnamese Machine Translation}
% \begin{tabular}{|m{1.5cm}|m{3.5cm}|m{4cm}|m{1.5cm}|m{3cm}|}
% \toprule
% \textbf{Team} & \textbf{Methodology} & \textbf{Model} & \textbf{Direction} & \textbf{Desc} \\
% \midrule
% BGSV AI & Using a sequence-to-sequence architecture leveraging a custom T5 model ~\cite{Raffel:2020} developed a custom tokenizer from scratch, employing the Sentence Piece technique, and combined it with our custom T5 architecture for this task Model. & Develop a new tokenizer set based on the SentencePiece ~\cite{kudo-richardson-2018-sentencepiece} technique to create a distinct vocabulary set adapted the T5 language model ~\cite{Raffel:2020} to train it from scratch for the machine translation task & lo-vi \newline vi-lo &  \\
% \bottomrule
% \end{tabular}
% %\label{tb5}
% \end{center}

The system illustrated in the Figure \ref{fig6:flow6} represents a customized T5 model developed by the team for bidirectional translation between Lao and Vietnamese. They created separate tokenizers for both languages using SentencePiece, while also performing preprocessing and adjusting token lengths as appropriate. As a result, the model is trained from scratch to optimize machine translation for both Lao and Vietnamese.

\section{Experimental Results}
\label{sec:experiment}
For all language pairs, we show the case-sensitive BLEU and SacreBLEU scores. Results would be ranked by human evaluation.
\begin{itemize}
    \item Only constrained systems will be evaluated and ranked.
    \item Unconstrained systems would not be human-evaluated and ranked.
    \item By Human: expert in Chinese and Lao languages (05 expert)
\end{itemize}

\textbf{VLSP 2022-MT}

%For all language pairs but one, we show the case-sensitive BLEU and SacreBLEU scores. The exception is the Vietnamese to Chinese task, for which character-level scores are given.
We observed that all participant systems outperformed the baselines, with tasks involving Chinese and Vietnamese attracting particular attention. For Chinese language, which is a language notoriously difficult to process, the better systems largely beat the basic methods featured in the baselines. For Vietnamese language, participant scores vary a lot as well; differently than on Chinese, submitted runs hardly provided higher quality than baselines; in particular, on Vietnamese-to-Chinese direction, none was able to improve the baseline translation: despite a deep analysis, we were unable to find a plausible explanation for this surprising outcome. The table~\ref{tb:vlsp2022method} shows the methodology for the teams in VLSP 2022 MT shared-task Chinese-Vietnamese.

\begin{table}[]
\centering
\caption{The summary of methodology for MT shared-task Chinese-Vietnamese}
\label{tb:vlsp2022method}
\begin{tabular}{|l|>{\centering\arraybackslash}m{2cm}|p{8cm}|}
\hline
No & Teams         & Methodology                                                                                                                                                \\ \hline
1  & VBD-MT        & Pre-trained model mBART, using sampling method for backtranslation, applying ensembling and postprocessing to improve the translation quality.         \\ \hline
2  & SDS           & Pre-trained language model mBART for the machine translation task, proposing data selection and data synthesis techniques from the monolingual corpus. \\ \hline
3  & S-NLP         & No technical report                                                                                                                                           \\ \hline
4  & VC-datamining & Using transformer model and integrating: data filtering, checkpoint averaging, data augmentation, ensemble.                                            \\ \hline
5  & JNLP          & Using Phrase Transformer for incorporating the phrase dependencies information into the Self-Attention mechanism.                                       \\ \hline
\end{tabular}
\end{table}

In table \ref{tb:vizhauto}, the Chinese-Vietnamese translation direction, the top three teams significantly outperformed the remaining two teams. Although the S-NLP team, ranked third, scored much lower in BLEU compared to the top two teams, they achieved a substantially higher score in ScareBLEU.

\begin{table}[H]
    \centering
    \caption{Vietnamese to Chinese Machine Translation Task (Automatic Evaluation Results)}
    \label{tb:vizhauto}
    \begin{tabularx}{0.65\textwidth} {
    | >{\centering\arraybackslash}c
    | >{\centering\arraybackslash}X
    | >{\centering\arraybackslash}X
    | >{\centering\arraybackslash}X|
    }
    \hline
    \bfseries No & \bfseries Team Name  & \bfseries BLEU & \bfseries SacreBLEU \\ \hline
        1 & S-NLP & 26.62 & 26.65 \\ \hline
        2 & SDS & 21.87 & 21.85 \\ \hline
        3 & JNLP & 21.70 & 21.76 \\ \hline
        4 & VBD-MT & 17.95 & 18.02 \\ \hline
        5 & VC-datamining & 17.10 & 17.15 \\ \hline
    \end{tabularx}
        
\end{table}

In table \ref{tb:vizhauto} for the Vietnamese-Chinese translation direction, the S-NLP team performed exceptionally well, surpassing the second-ranked team by nearly 5 points in both BLEU and ScareBLEU scores.

\textbf{VLSP 2023-MT} 

The teams engaged in the Vietnamese-Lao machine translation task with the aim of enhancing the quality of translation models for this language pair. The methods implemented by the teams significantly outperformed baseline models on the test dataset, especially under human evaluation. The summary of methodology machine translation in VLSP 2023 for Lao-Viet language pair given by in the table~\ref{tb:vlsp2023method}.

\begin{table}[]
\centering
\caption{The summary of methodology for MT shared-task Lao-Vietnamese}
\label{tb:vlsp2023method}
\begin{tabular}{|l|>{\centering\arraybackslash}m{2cm}|p{8cm}|}
\hline
\textbf{No} & \textbf{Teams} & \textbf{Methodology}                                                                                                                                                                                            \\ \hline
1           & BGSV AI      & Using pre-trained model T5 and Sentence Piece to create a distinct vocabulary set adapted the T5 to train it from scratch for the machine translation task                                                \\ \hline
2           & TESTLAV100     & Using OpenNMT and Sentencepiece to tokenizer, apply backtranslation and model weight averaging to optimize performance.                                                                                      \\ \hline
3           & FAIZ AIO     & Using Transformer wmt en de big model for Vietnamese to Lao machine translation                                                                                                                               \\ \hline
4           & BLUESKY        & Pre-trained mBART model based on monolingual Vietnamese and Lao for MT task.                                                                                                                                  \\ \hline
5           & MTA AI       & Using pre-trained models T5, m2m100, and effective backtranslation methods to address the limitations of the Vietnamese-Lao bilingual data. Using M2M-100 418M model and mt5-small for finetuning system \\ \hline
6           & HUMBLE BEES   & No technical report                                                                                                                                                                                           \\ \hline
7           & TTS66          & Using transformer models, experiment with recurrent neural networks (RNN), optimizing hyper-parameters, and tagged backtranslation                               \\ \hline
\end{tabular}
\end{table}

\begin{table}[H]
    \centering
    \caption{Lao to Vietnamese Machine Translation Task (Automatic Evaluation Results)}
    \label{tb:loviauto}
    \begin{tabularx}{0.85\textwidth} {
    | c % No column
    | >{\centering\arraybackslash}X % Team Name auto-adjusted width
    | c % SacreBLEU
    | c % Human
    | c % FinalScore
    |}
    \hline
    \textbf{No} & \textbf{Team Name} & \textbf{SacreBLEU} & \textbf{Human} & \textbf{FinalScore} \\ \hline
        1 & BGSV AI & 32.56 & 27.51 & 27.51 \\ \hline
        2 & TESTLAV100 & 9.92 & 12.05 & 12.05 \\ \hline
        3 & FAIZ AIO & 42.19 & 47.83 & 47.83 \\ \hline
        4 & BLUESKY & 49.67 & 54.28 & 54.28 \\ \hline
        5 & HUMBLE BEES & 28.45 & 26.94 & 26.94 \\ \hline
        6 & TTS66 & 17.92 & 33.73 & 33.73 \\ \hline
        7 & MTA AI & 26.03 & 51.03 & 51.03 \\ \hline
    \end{tabularx}
\end{table}

In table \ref{tb:loviauto} with the Lao-Vietnamese translation direction, three teams achieved significantly higher scores compared to the others. Notably, the human evaluation scores for these teams were exceptionally high, playing a decisive role in determining their rankings. 

\begin{table}[H]
    \centering
    \caption{Vietnamese to Lao Machine Translation Task (Automatic Evaluation Results)}
    \label{tb:viloauto}
    \begin{tabularx}{0.85\textwidth} {
    | >{\centering\arraybackslash}c % No column
    | >{\centering\arraybackslash}X % Team Name auto-adjusted
    | >{\centering\arraybackslash}c % SacreBLEU
    | >{\centering\arraybackslash}c % Human
    | >{\centering\arraybackslash}c % FinalScore
    |}
    \hline
    \bfseries No & \bfseries Team Name  & \bfseries SacreBLEU & \bfseries Human & \bfseries FinalScore \\ \hline
        1 & BGSV AI & 29.21 & 31.56 & 31.56 \\ \hline
        2 & TESTLAV100 & 28.09 & 20.41 & 20.41 \\ \hline
        3 & FAIZ\_AIO & 38.04 & 46.51 & 46.51 \\ \hline
        4 & BLUESKY & 43.08 & 51.37 & 51.37 \\ \hline
        5 & MTA\_AI & 41.88 & 61.31 & 61.31 \\ \hline
    \end{tabularx}
\end{table}

In table~\ref{tb:loviauto} the Viet-Lao translation direction, two teams achieved outstanding results compared to the others, with both scoring over 50 points. Notably, the MTA\_AI team achieved a score of 61. In the final results, the human evaluation scores were decisive in determining the rankings. Although the MTA\_AI team had a lower ScareBLEU score, their human evaluation score was exceptional.
\section{Human Evaluation}
To accurately and comprehensively evaluate the quality of the translation model, we decided to leverage the expertise and experience of specialists in the field of linguistics. The evaluation process began by obtaining sentence translations from various models and then submitting these translations to experts for review and scoring. This process went beyond merely comparing results with standard translations; it required experts to analyze and assess based on multiple factors such as semantic accuracy, syntax, context, and the naturalness of the translated language. With their deep understanding of language and grammar, the experts provided feedback and evaluations that closely reflect everyday language use. Employing human evaluation in this manner offers us a proactive and insightful perspective on the model's actual performance, ensuring that the final results are not only technically accurate but also appropriate and easily understandable for users.

\textit{VLSP 2022- MT Chinese-Vietnamese:}
The human evaluation process was conducted on translations generated by the models. These tasks included translating from Vietnamese to Chinese (Vi-Zh) and from Chinese to Vietnamese (Zh-Vi). During this evaluation, translations produced by the models were reviewed and assessed by linguistic experts with extensive skills and knowledge in both languages involved. The objective was to determine the quality, accuracy, and naturalness of the translations to evaluate the performance of the machine translation models.

\textit{VLSP 2023-MT Lao-Vietnamese:} Human evaluation was carried out on primary runs submitted by participants to two of the MT tasks, namely the MT Vietnamese-Lao (Vi-Lo) task and MT Lao -Vietnamese (Lo-Vi) task. 

From the point of view of the evaluation campaign, our goal is to adopt a human evaluation framework able to maximize the benefit for the research community, both in terms of information about MT systems and data and resources to be reused. With respect to other types of human assessment, such as judgments of translation quality (i.e. adequacy/fluency and ranking tasks), the post-editing task has the double advantage of producing (i) a set of edits pointing to specific
translation errors, and (ii) a set of additional reference translations. Both these byproducts are very useful for MT system development and evaluation. The human evaluation dataset and the collected post-edits are described in next Section whereas the results of the evaluation are presented in result table.

%\subsection{Evaluation Data}
\textbf{\textit{Evaluation Dataset}}
%\textbf{VLSP 2022-MT Chinese $\longleftrightarrow$ Vietnamese and VLSP 2023-MT Lao $\longleftrightarrow$ Vietnamese:} 

The human evaluation datasets each consist of approximately 1,000 sentences, drawn from subsets of the private test sets for each translation task. Specifically, we selected 1,000 sentences for the Zh-Vi and Vi-Zh datasets, and another 1,000 sentences for the Lo-Vi and Vi-Lo datasets. This approach, which involves selecting a consecutive block of sentences for each dataset, was guided by the need to realistically simulate a caption post-editing task.

We received five submissions for each of the Zh-Vi, Vi-Zh, Vi-Lo, and Lo-Vi tasks. For each task, the output from the five systems was given to five professional translators for post-editing on the human evaluation set. 
% \textbf{VLSP 2023-MT Lao $\longleftrightarrow$ Vietnamese:} The human evaluation datasets contain around 1,000 sentences each and include subsets of the
% private test sets. We selected 100\% of each talk for the Lo-Vi dataset and Vi-Lo one. This choice
% of selecting a consecutive block of sentences for each talk was determined by the need of
% realistically simulating a caption post-editing task on several. 
% We received five submissions both for the Lo-Vi task and the Vi-Lo task. For Lo-Vi task, the output of the seven systems on the human evaluation set was assigned to five professional translators to be post-edited. For Vi-Lo task, the output of the five systems on the human evaluation
To cope with translators’ variability, an equal number of outputs from each MT system was assigned randomly to each translator. The resulting evaluation data for each task consist of the new reference translations for each of the sentences in the human evaluation set. Each one of these references represents the targeted translation of the system output from which it was derived, and remain additional translations are available as well for the evaluation of each MT system.

\begin{figure}[h]
    \centering
    \includegraphics[width=16cm, height=6cm]{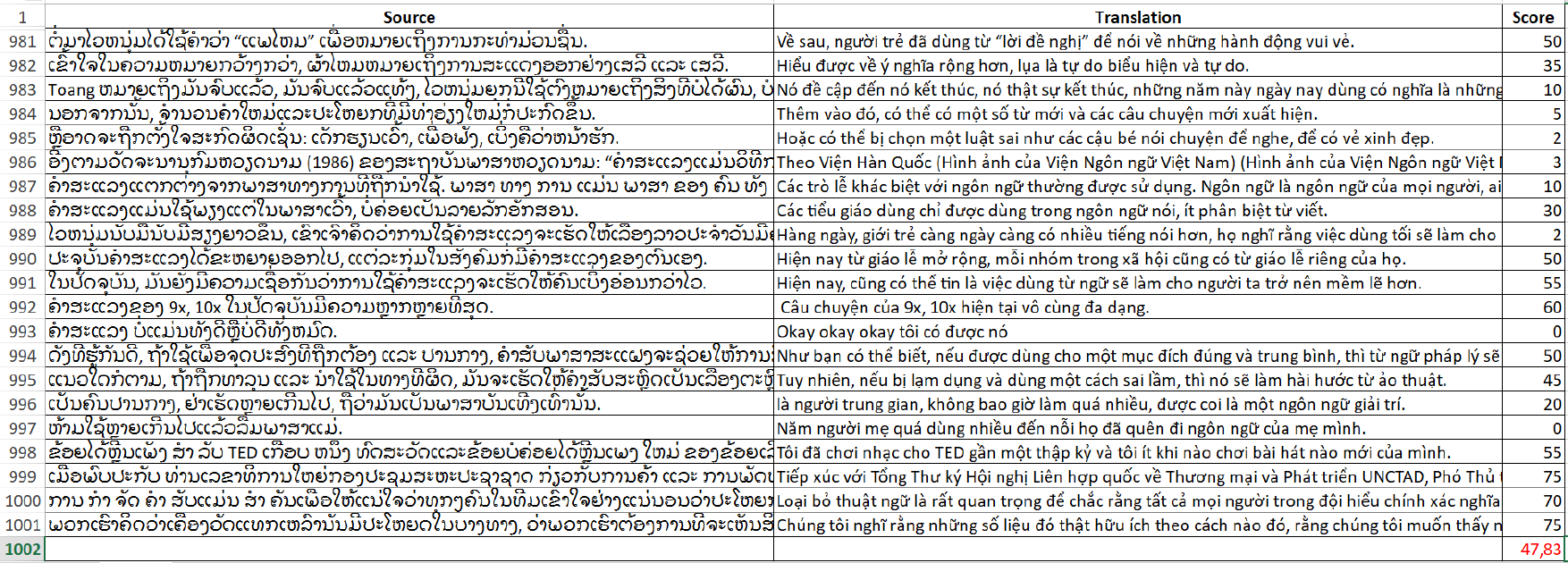}
    \caption {An example evaluation by human for Lao - Vietnamese machine translation task}
    \label{image1}
\end{figure}
%\subsection{Result}

\textbf{\textit{Human evaluation results}}

The outcomes of the two previous rounds of human evaluation through post-editing demonstrated that human evaluation error computed against all the references produced by all post-editors allow a more reliable and consistent evaluation of MT systems with respect to human evaluation error calculated against the targeted reference only. In light of these findings, also this year systems were officially ranked according to human evaluation error calculated on all the collected postedits. In figure \ref{image1} shows the example evaluation by human for Lao - Vietnamese machine translation task (evaluation with the expert in Lao language).
% \begin{table}[H]
% \centering
% \caption{Human evaluation results. Scores are given in percentage (\%). The system name next to the human evaluation error score indicates the first system in the ranking with which differences are statistically significant at \(p < 0.01\) (an asterisk indicates significance at \(p < 0.05\)).}
% \label{tb:4}
% \begin{tabular}{|l|l|l|l|}
% \hline
% \textbf{Task}            & \textbf{Rank} & \textbf{Team Name} & \textbf{Result} \\ \hline
% \multirow{5}{*}{Zh $\rightarrow$ Vi} & 1             & SDS                & 74.73           \\ \cline{2-4}
%                          & 2             & VBD-MT             & 71.42           \\ \cline{2-4}
%                          & 3             & S-NLP              & 68.74           \\ \cline{2-4}
%                          & 4             & JNLP               & 65.29           \\ \cline{2-4}
%                          & 5             & VC-Datamining      & 64.40           \\ \hline
% \multirow{5}{*}{Vi $\rightarrow$ Zh} & 1             & S-NLP              & 73.58           \\ \cline{2-4}
%                          & 2             & VBD-MT             & 69.19           \\ \cline{2-4}
%                          & 3             & VC-Datamining      & 67.80           \\ \cline{2-4}
%                          & 4             & SDS                & 67.68           \\ \cline{2-4}
%                          & 5             & JNLP               & 67.08           \\ \hline
% \end{tabular}
% \end{table}
% % \end{tabular}

\begin{table}[H]
\centering
\caption{Human evaluation results for Chinese-Vietnamese MT, scores are given in percentage (\%).} 
%The system name next to the human evaluation error score indicates the first system in the ranking with which differences are statistically significant at \(p < 0.01\) (an asterisk indicates significance at \(p < 0.05\)).}
\label{tb:result2022}

\begin{tabular}{|c|c|c|c|}
\hline
\textbf{Task}            & \textbf{Rank} & \textbf{Team Name} & \textbf{Result} \\ \hline
\multirow{5}{*}{Zh $\rightarrow$ Vi} & 1             & SDS                & 74.73           \\ \cline{2-4}
                         & 2             & VBD-MT             & 71.42           \\ \cline{2-4}
                         & 3             & S-NLP              & 68.74           \\ \cline{2-4}
                         & 4             & JNLP               & 65.29           \\ \cline{2-4}
                         & 5             & VC-Datamining      & 64.40           \\ \hline
\multirow{5}{*}{Vi $\rightarrow$ Zh} & 1             & S-NLP              & 73.58           \\ \cline{2-4}
                         & 2             & VBD-MT             & 69.19           \\ \cline{2-4}
                         & 3             & VC-Datamining      & 67.80           \\ \cline{2-4}
                         & 4             & SDS                & 67.68           \\ \cline{2-4}
                         & 5             & JNLP               & 67.08           \\ \hline
\end{tabular}

\end{table}

In VLSP 2022, the official results and rankings are presented in bold in Tables \ref{tb:result2022} and \ref{tb:rank2022}, which also present human evaluation error scores calculated on the targeted reference only and results, both on the human evaluation error set and on the full test set, calculated against the official reference translation used for automatic evaluation (see Section \ref{syssub}). Due to various reasons, the S-NLP team could not complete the technical report, so we have removed the S-NLP team from the final standings. As you can see in Table \ref{tb:result2022}, For the Vi $\rightarrow$ Zh task, the top-ranked system (VBD-MT) is significantly better than all the other systems, while VC-Datamining, SDS and JNLP are not different from each other. On the other hand, For the Zh $\rightarrow$ Vi task, SDS achieve the highest score, followed by VBDMT, JNLP, and VC-Datamining. Finally, we calculate the average score of both tasks to choose the champion team. The “Final Score” column of Table \ref{tb:rank2022} shows that the winning team is SDS, second is VBD-MT, third is JNLP, and fourth is VC-Datamining. However, there is no system that is significantly better than all other systems; the three top-ranking systems (SDS, VBD-MT, JNLP) are significantly better than the bottom-ranking systems (VC-datamining).

To conclude, the post-editing task introduced for manual evaluation brought benefit to the VLSP community, and in general to the MT field. Indeed, producing post-edited versions of the participating systems’ outputs allowed us to carry out a quite informative evaluation which minimizes the variability of post-editors, who naturally tend to diverge from the post-editing guidelines and personalize their translations. Furthermore, a number of additional reference translations are made available to the community for further development and evaluation of MT systems

% \begin{table}[H]
% \centering
% \caption{Final ranking results. Scores are given in percentage ($\%$)}
% \label{tb:5}
% \begin{tabular}{|c|l|c|c|c|}
% \hline
% \multicolumn{1}{|l|}{\textbf{Rank}} & \textbf{Team Name}                                       & \multicolumn{1}{l|}{\textbf{Zh $\rightarrow$Vi}} & \multicolumn{1}{l|}{\textbf{Vi $\rightarrow$ Zh}} & \multicolumn{1}{l|}{\textbf{Final Score}} \\ \hline
% 1                                   & SDS                                                      & 74,73                                                & 67,68                                 & 71,27                                     \\ \hline
% 2                                   & VBD-MT                                                   & 71,42                                                & 69,19                                 & 70,30                                     \\ \hline
% 3                                   & JNLP                                                     & 65,29                                                & 67,08                                 & 66,18                                     \\ \hline
% 4                                   & \begin{tabular}[c]{@{}l@{}} VC-Datamining\end{tabular} & 64,40                                                & 67,80                                 & 65,60                                     \\ \hline
% \end{tabular}
% \end{table}
\begin{table}[H]
\centering
\caption{Final ranking results for Chinese-Vietnamese MT. Scores are given in percentage ($\%$)}
\label{tb:rank2022}
\begin{tabular}{|c|c|c|c|c|}
\hline
\textbf{Rank} & \textbf{Team Name} & \textbf{Zh $\rightarrow$ Vi} & \textbf{Vi $\rightarrow$ Zh} & \textbf{Final Score} \\ \hline
1 & SDS & 74.73 & 67.68 & 71.27 \\ \hline
2 & VBD-MT & 71.42 & 69.19 & 70.30 \\ \hline
3 & JNLP & 65.29 & 67.08 & 66.18 \\ \hline
4 & \begin{tabular}[c]{@{}c@{}} VC-Datamining \end{tabular} & 64.40 & 67.80 & 65.60 \\ \hline
\end{tabular}
\end{table}

In VLSP 2023: Official results and rankings are presented in bold in Tables \ref{tab:competition_results}, which also present human evaluation error scores calculated on the targeted reference only and results, both on the human evaluation error set and on the full test set, calculated against the official reference translation used for automatic evaluation. Due to various reasons, the S-NLP team could not complete the technical report, so we have removed the S-NLP team from the final standings. As you can see in Table \ref{tab:competition_results}, for the Vi $ \rightarrow $ Lo task, the top-ranked system (MTA\_AI ) is significantly better than all the other systems (Bluesky, Faiz\_AIO and BGSV\_AI). On the other hand, for the Lo $ \rightarrow $ Vi task, Bluesky achieve the highest score, followed by MTA\_AI, Faiz\_AIO and BGSV\_AI.

Finally, we calculate the average score of both tasks to choose the champion team. The “Final Score” column of Table \ref{tab:competition_results} shows that the winning team is Bluesky, second is MTA\_AI, third is BGSV\_AI).
% \begin{table}[H]
% \centering
% \caption{Final ranking results for Lao $\leftrightarrow$ Vietnamese MT. Scores are given in percentage (\%).}
% \label{tab:competition_results}
% \begin{tabular}{|l|l|l|l|l|}
% \hline
% \textbf{Rank} & \textbf{Team Name} & \textbf{lo-vi} & \textbf{vi-lo} & \textbf{Description} \\ \hline
% \multirow{2}{*}{1} & Bluesky   & 54.28   & 51.37   & SacreBLEU highest \\ \cline{2-5}
%                    & MTA\_AI   & 51.03   & 61.31   &                   \\ \hline
% \multirow{2}{*}{2} & Faiz\_AIO & 47.83   & 46.51   & No technical report \\ \cline{2-5}
%                    & BGSV\_AI  & 27.51   & 31.56   &                   \\ \hline
% \multirow{3}{*}{3} & TTS66     & 33.73   & -       &                   \\ \cline{2-5}
%                    & Humble Bees & 26.94   & -       &                   \\ \cline{2-5}
%                    & TeslaV100 & 12.05   & 20.41   &                   \\ \hline
% \end{tabular}
% \end{table}

\begin{table}[H]
\centering
\caption{Final ranking results for Lao $\leftrightarrow$ Vietnamese MT. Scores are given in percentage (\%).}
\label{tab:competition_results}
\begin{tabular}{|c|c|c|c|c|}
\hline
\textbf{Rank} & \textbf{Team Name} & \textbf{Lo-Vi} & \textbf{Vi-Lo} & \textbf{Description} \\ \hline
\multirow{2}{*}{1} & Bluesky   & \textbf{54.28} & 51.37   & SacreBLEU highest \\ \cline{2-5}
                   & MTA\_AI   & 51.03   & \textbf{61.31} &                   \\ \hline
\multirow{2}{*}{2} & Faiz\_AIO & 47.83   & 46.51   & No technical report \\ \cline{2-5}
                   & BGSV\_AI  & 27.51   & 31.56   &                   \\ \hline
\multirow{3}{*}{3} & TTS66     & 33.73   & N/A       &                   \\ \cline{2-5}
                   & Humble Bees & 26.94   & N/A       &                   \\ \cline{2-5}
                   & TeslaV100 & 12.05   & 20.41   &                   \\ \hline
\end{tabular}

\end{table}
\section{Conclusions}

In this paper, we presented the organization and outcomes of the VLSP MT Evaluation Campaign.
The VLSP MT evaluation provides a venue where core technologies for spoken language translation can be evaluated on many different languages and compared not only across research teams but also overtime.

\begin{itemize}
    \item In VLSP 2022, the evaluation was attended by 5 groups: Samsung SDS R\&D Center, Vin BigData, Japan Advanced Institute of Science and Technology, Hanoi University of Science and Technology, and VCCorp.
    \item In VLSP 2023, the evaluation was attended by 7 groups: UET-VNU, MTA, Viettel, Bosch Global Software Technologies Vietnam, HUST, Fulbright University Vietnam, US-VNUHCM. To honor the local organizer, we added among the offered translation directions also Vietnamese-Chinese and Vietnamese-Lao, which finally attracted several participants.
\end{itemize}

% Finally, for the most popular MT runs, a manual evaluation was carried out with professional translators aiming at measuring MT quality in terms of post-editing effort required to fix the MT outputs. In future plans, we are considering to extend the translation task using Pre-train models and Vietnamese Large Language Models to other languages: Chinese, Lao, Khmer

Finally, to assess the quality of the most successful machine translation runs, a manual evaluation was conducted by professional translators. The goal was to determine the amount of post-editing required to correct the machine-generated translations. Looking ahead, we aim to expand the translation task by incorporating pre-trained models and Vietnamese large language models to support additional languages, such as Chinese, Lao, and Khmer.

\section*{Acknowledgments}

% The human evaluation and part of the work by the Ministry of Science and Technology of Vietnam
% under Program KC 4.0, No. KC-4.0.12/19-25. We would like to express to the organizers of the
% Vietnamese Language and Speech Processing (VLSP) for their exceptional dedication in
% orchestrating the Machine Translation challenge. Their efforts have been instrumental in fostering advancements in the field. We also would like to express our sincere gratitude to the members from the NLP-UET Lab, University of Engineering and Technology -Vietnam National
% University, Hanoi, Vietnam. They have contributed to the completion of this research

This work is part of the Ministry of Science and Technology of Vietnam's Program KC 4.0, project No. KC-4.0.12/19-25. We would like to extend our appreciation to the Vietnamese Language and Speech Processing (VLSP) organizers for their exceptional commitment in coordinating the Machine Translation challenge, which has significantly contributed to progress in this field. We are also deeply grateful to the team from the NLP-UET Lab at the University of Engineering and Technology, Vietnam National University, Hanoi, for their invaluable contributions to this research.
\printbibliography

@article{Papineni,
author = {Papineni, Kishore and Roukos, Salim and Ward, Todd and Zhu, Wei Jing},
year = {2002},
month = {10},
pages = {},
title = {BLEU: a Method for Automatic Evaluation of Machine Translation},
doi = {10.3115/1073083.1073135}
}

@article{Matt:2018,
    title = {A Call for Clarity in Reporting {BLEU} Scores},
    author = {Post, Matt},
    booktitle = {Proceedings of the Third Conference on Machine Translation: Research Papers},
    month = {10},
    year = {2018},
    address = {Brussels, Belgium},
    publisher = {Association for Computational Linguistics},
    url = {https://aclanthology.org/W18-6319},
    doi = {10.18653/v1/W18-6319},
    pages = {186--191}
}

@inproceedings{Cho2014LearningPR,
  title={Learning Phrase Representations using RNN Encoder–Decoder for Statistical Machine Translation},
  author={Kyunghyun Cho and Bart van Merrienboer and Çaglar G{\"u}lçehre and Dzmitry Bahdanau and Fethi Bougares and Holger Schwenk and Yoshua Bengio},
  booktitle={Conference on Empirical Methods in Natural Language Processing},
  year={2014},
  url={https://api.semanticscholar.org/CorpusID:5590763}
}

@inproceedings{Sutskever:2014,
  title={Sequence to Sequence Learning with Neural Networks},
  author={Ilya Sutskever and Oriol Vinyals and Quoc V. Le},
  booktitle={NIPS},
  year={2014}
}

@article{Vaswani:2017,
  author    = {Ashish Vaswani and
               Noam Shazeer and
               Niki Parmar and
               Jakob Uszkoreit and
               Llion Jones and
               Aidan N. Gomez and
               Lukasz Kaiser and
               Illia Polosukhin},
  title     = {Attention Is All You Need},
  journal   = {CoRR},
  volume    = {abs/1706.03762},
  year      = {2017},
  url       = {http://arxiv.org/abs/1706.03762}
}

@article{Wu:2016,
  author    = {Yonghui Wu and
               Mike Schuster and
               Zhifeng Chen and
               Quoc V. Le and
               Mohammad Norouzi},
  title     = {Google\'s Neural Machine Translation System: Bridging the Gap between Human and Machine Translation},
  journal   = {CoRR},
  volume    = {abs/1609.08144},
  year      = {2016},
  url       = {http://arxiv.org/abs/1609.08144}
}

@article{KleinKDSR17,
  author    = {Guillaume Klein and
               Yoon Kim and
               Yuntian Deng and
               Jean Senellart and
               Alexander M. Rush},
  title     = {OpenNMT: Open-Source Toolkit for Neural Machine Translation},
  journal   = {CoRR},
  volume    = {abs/1701.02810},
  year      = {2017},
  url       = {http://arxiv.org/abs/1701.02810}
}

@inproceedings{koehn:2017,
    title = {Six Challenges for Neural Machine Translation},
    author = {Koehn, Philipp  and
      Knowles, Rebecca},
    booktitle = {Proceedings of the First Workshop on Neural Machine Translation},
    month = {8},
    year = {2017},
    address = {Vancouver},
    publisher = {Association for Computational Linguistics},
    url = {https://aclanthology.org/W17-3204},
    doi = {10.18653/v1/W17-3204},
    pages = {28--39}
}

@article{NIST:2008,
author = {Przybocki, Mark and Peterson, Kay and Bronsart, Sebastien and Sanders, Gregory},
year = {2009},
month = {09},
pages = {71-103},
title = {The NIST 2008 Metrics for machine translation challenge—overview, methodology, metrics, and results},
volume = {23},
journal = {Machine Translation},
doi = {10.1007/s10590-009-9065-6}
}

@inproceedings{TER:2008,
  title={Meteor, M-BLEU and M-TER: Evaluation Metrics for High-Correlation with Human Rankings of Machine Translation Output},
  author={Abhaya Agarwal and Alon Lavie},
  booktitle={WMT@ACL},
  year={2008}
}

@inproceedings{ott-etal-2018-scaling,
    title = "Scaling Neural Machine Translation",
    author = "Ott, Myle  and
      Edunov, Sergey  and
      Grangier, David  and
      Auli, Michael",
    editor = "Bojar, Ond{\v{r}}ej  and
      Chatterjee, Rajen  and
      Federmann, Christian  and
      Fishel, Mark  and
      Graham, Yvette  and
      Haddow, Barry  and
      Huck, Matthias  and
      Yepes, Antonio Jimeno  and
      Koehn, Philipp  and
      Monz, Christof  and
      Negri, Matteo  and
      N{\'e}v{\'e}ol, Aur{\'e}lie  and
      Neves, Mariana  and
      Post, Matt  and
      Specia, Lucia  and
      Turchi, Marco  and
      Verspoor, Karin",
    booktitle = "Proceedings of the Third Conference on Machine Translation: Research Papers",
    month = oct,
    year = "2018",
    address = "Brussels, Belgium",
    publisher = "Association for Computational Linguistics",
    url = "https://aclanthology.org/W18-6301",
    doi = "10.18653/v1/W18-6301",
    pages = "1--9",
}

@inproceedings{xue-etal-2021-mt5,
    title = "m{T}5: A Massively Multilingual Pre-trained Text-to-Text Transformer",
    author = "Xue, Linting  and
      Constant, Noah  and
      Roberts, Adam  and
      Kale, Mihir  and
      Al-Rfou, Rami  and
      Siddhant, Aditya  and
      Barua, Aditya  and
      Raffel, Colin",
    editor = "Toutanova, Kristina  and
      Rumshisky, Anna  and
      Zettlemoyer, Luke  and
      Hakkani-Tur, Dilek  and
      Beltagy, Iz  and
      Bethard, Steven  and
      Cotterell, Ryan  and
      Chakraborty, Tanmoy  and
      Zhou, Yichao",
    booktitle = "Proceedings of the 2021 Conference of the North American Chapter of the Association for Computational Linguistics: Human Language Technologies",
    month = jun,
    year = "2021",
    address = "Online",
    publisher = "Association for Computational Linguistics",
    url = "https://aclanthology.org/2021.naacl-main.41",
    doi = "10.18653/v1/2021.naacl-main.41",
    pages = "483--498",
}

@misc{kingma2017adam,
      title={Adam: A Method for Stochastic Optimization}, 
      author={Diederik P. Kingma and Jimmy Ba},
      year={2017},
      eprint={1412.6980},
      archivePrefix={arXiv},
      primaryClass={cs.LG}
}

@misc{xue2021mt5,
      title={mT5: A massively multilingual pre-trained text-to-text transformer}, 
      author={Linting Xue and Noah Constant and Adam Roberts and Mihir Kale and Rami Al-Rfou and Aditya Siddhant and Aditya Barua and Colin Raffel},
      year={2021},
      eprint={2010.11934},
      archivePrefix={arXiv},
      primaryClass={cs.CL}
}

@misc{fan2020englishcentric,
      title={Beyond English-Centric Multilingual Machine Translation}, 
      author={Angela Fan and Shruti Bhosale and Holger Schwenk and Zhiyi Ma and Ahmed El-Kishky and Siddharth Goyal and Mandeep Baines and Onur Celebi and Guillaume Wenzek and Vishrav Chaudhary and Naman Goyal and Tom Birch and Vitaliy Liptchinsky and Sergey Edunov and Edouard Grave and Michael Auli and Armand Joulin},
      year={2020},
      eprint={2010.11125},
      archivePrefix={arXiv},
      primaryClass={cs.CL}
}

@article{Raffel:2020,
  author  = {Colin Raffel and Noam Shazeer and Adam Roberts and Katherine Lee and Sharan Narang and Michael Matena and Yanqi Zhou and Wei Li and Peter J. Liu},
  title   = {Exploring the Limits of Transfer Learning with a Unified Text-to-Text Transformer},
  journal = {Journal of Machine Learning Research},
  year    = {2020},
  volume  = {21},
  number  = {140},
  pages   = {1--67},
  url     = {http://jmlr.org/papers/v21/20-074.html}
}

@inproceedings{kudo-richardson-2018-sentencepiece,
    title = "{S}entence{P}iece: A simple and language independent subword tokenizer and detokenizer for Neural Text Processing",
    author = "Kudo, Taku  and
      Richardson, John",
    editor = "Blanco, Eduardo  and
      Lu, Wei",
    booktitle = "Proceedings of the 2018 Conference on Empirical Methods in Natural Language Processing: System Demonstrations",
    month = nov,
    year = "2018",
    address = "Brussels, Belgium",
    publisher = "Association for Computational Linguistics",
    url = "https://aclanthology.org/D18-2012",
    doi = "10.18653/v1/D18-2012",
    pages = "66--71",
}

@inproceedings{ott-etal-2019-fairseq,
    title = "fairseq: A Fast, Extensible Toolkit for Sequence Modeling",
    author = "Ott, Myle  and
      Edunov, Sergey  and
      Baevski, Alexei  and
      Fan, Angela  and
      Gross, Sam  and
      Ng, Nathan  and
      Grangier, David  and
      Auli, Michael",
    editor = "Ammar, Waleed  and
      Louis, Annie  and
      Mostafazadeh, Nasrin",
    booktitle = "Proceedings of the 2019 Conference of the North {A}merican Chapter of the Association for Computational Linguistics (Demonstrations)",
    month = jun,
    year = "2019",
    address = "Minneapolis, Minnesota",
    publisher = "Association for Computational Linguistics",
    url = "https://aclanthology.org/N19-4009",
    doi = "10.18653/v1/N19-4009",
    pages = "48--53",
}

@article{Sennrich2015NeuralMT,
  title={Neural Machine Translation of Rare Words with Subword Units},
  author={Rico Sennrich and Barry Haddow and Alexandra Birch},
  journal={ArXiv},
  year={2015},
  volume={abs/1508.07909},
  url={https://api.semanticscholar.org/CorpusID:1114678}
}

@article{Tang2020MultilingualTW,
  title={Multilingual Translation with Extensible Multilingual Pretraining and Finetuning},
  author={Y. Tang and C. Tran and Xian Li and Peng-Jen Chen and Naman Goyal and Vishrav Chaudhary and Jiatao Gu and Angela Fan},
  journal={ArXiv},
  year={2020},
  volume={abs/2008.00401},
  url={https://api.semanticscholar.org/CorpusID:220936592}
}
%\printbibliography %Prints bibliography
\end{document}